\documentclass[10pt,twocolumn,letterpaper]{article}

\usepackage{iccv}
\usepackage{times}
\usepackage{epsfig}
\usepackage{graphicx, subfigure}
\usepackage{amsmath}
\usepackage{amssymb}
\usepackage{multirow}
\usepackage{makecell}
\usepackage{authblk}
\usepackage{pifont}
\newcommand{\cmark}{\ding{51}}
\newcommand{\xmark}{\ding{55}}

\usepackage{setspace}

\usepackage[pagebackref=true,breaklinks=true,letterpaper=true,colorlinks,bookmarks=false]{hyperref}
\usepackage{amsmath,amsfonts,bm}
\def\eqref#1{equation~\ref{#1}}
\def\1{\bm{1}}

\def\vc{{\bm{c}}}

\def\ve{{\bm{e}}}
\def\vf{{\bm{f}}}

\def\vo{{\bm{o}}}

\def\vs{{\bm{s}}}

\def\vw{{\bm{w}}}

\def\vz{{\bm{z}}}

\def\evw{{w}}

\def\mC{{\bm{C}}}

\def\mE{{\bm{E}}}

\DeclareMathAlphabet{\mathsfit}{\encodingdefault}{\sfdefault}{m}{sl}
\SetMathAlphabet{\mathsfit}{bold}{\encodingdefault}{\sfdefault}{bx}{n}
\newcommand{\tens}[1]{\bm{\mathsfit{#1}}}

\def\ttau{{\tens{\tau}}}

\def\gE{{\mathcal{E}}}

\def\gG{{\mathcal{G}}}

\def\gV{{\mathcal{V}}}

\newcommand{\sigmoid}{\sigma}

\iccvfinalcopy % *** Uncomment this line for the final submission

\def\iccvPaperID{8059} % *** Enter the ICCV Paper ID here

\begin{document}
%%%%%%%%% TITLE
\title{Progressive Correspondence Pruning by Consensus Learning}
\author{
Chen Zhao$^{1\dagger}$\thanks{Work was done when the author was an intern at SenseTime Research} \quad Yixiao Ge$^3$\thanks{Authors contributed equally} \quad Feng Zhu$^2$ \quad Rui Zhao$^{24}$ \quad Hongsheng Li$^3$ \quad Mathieu Salzmann$^1$ \\
$^1$École Polytechnique Fédérale de Lausanne (EPFL) \\
$^2$SenseTime Research $ $ $^3$The Chinese University of Hong Kong \\
$^4$Qing Yuan Research Institute, Shanghai Jiao Tong University \\
{\tt\small \{chen.zhao, mathieu.salzmann\}@epfl.ch} 
{\tt\small \{zhufeng, zhaorui\}@sensetime.com} \\
{\tt\small \{yxge@link, hsli@ee\}.cuhk.edu.hk}
}
\maketitle
\pagestyle{empty}  % no page number for the second and the later pages
\thispagestyle{empty} % no page number for the first page
%%%%%%%%% ABSTRACT
\begin{abstract}
%   Given a set of noisy feature correspondences from two views, recognizing the consistent ones is critical for high-accuracy reconstruction and localization. Some learning-based methods have been presented in the literature which predict binary labels (inlier or outlier) for initial correspondences. However, since mismatches are randomly distributed and inconsistent with each other, existing methods are vulnerable in the context of outliers taking a dominated part.  To address this issue, we introduce a hierarchical network which removes outliers via local-to-global consensus learning. Specifically, noisy correspondences are hierarchically down-sampled into a subset of candidates with fewer outliers. The local-to-global consensus is represented by dynamic graph-based learning, which facilitates the reliability of candidates. A parametric model is then estimated on the candidates and used to recognize consistent correspondences from the initial set. Comprehensively experiments are performed on several public datasets, which have shown that the performance of our method remarkably outperforms the state-of-the-art methods.
Correspondence selection aims to correctly select the consistent matches (inliers) from an initial set of putative correspondences. The selection is challenging since putative matches are typically extremely unbalanced, largely dominated by outliers, and the random distribution of such outliers further complicates the learning process for learning-based methods. To address this issue, we propose to progressively prune the correspondences via a local-to-global consensus learning procedure. We introduce a ``pruning'' block that lets us identify reliable candidates among the initial matches according to consensus scores estimated using local-to-global dynamic graphs. We then achieve progressive pruning by stacking multiple pruning blocks sequentially. Our method outperforms state-of-the-arts on robust line fitting, camera pose estimation and retrieval-based image localization benchmarks by significant margins and shows promising generalization ability to different datasets and detector/descriptor combinations.
\end{abstract}
\vspace{-10pt}
%%%%%%%%% BODY TEXT
\section{Introduction}
% Pixel-wise correspondence is pivotal to many tasks in computer vision and robotics, including structure from motion \textcolor{red}{(add reference)}, simultaneous location and mapping, image stitching, visual localization, virtual reality, and etc. Ideally, given a set of feature correspondences established by some off-the-shelf feature detectors and descriptors, the matched keypoints are supposed to represent consistent visual patterns, i.e., projections from different views of the same 3d point in real world. However, the mismatches are inevitable in practice due to the effect of various nuisances such as rotation, translation, scale change, viewpoint change, illumination change, and etc. To achieve high-accuracy feature matching, correspondence selection has been a critical post-process to select correct matches (inliers) and reject false matches (outliers). 
%correspondence selection has been widely used as a post processing to select correct matches (inliers) and reject false matches (outliers). 
\begin{figure}[!t]
    \begin{center}
        {\includegraphics[width=0.9\linewidth]{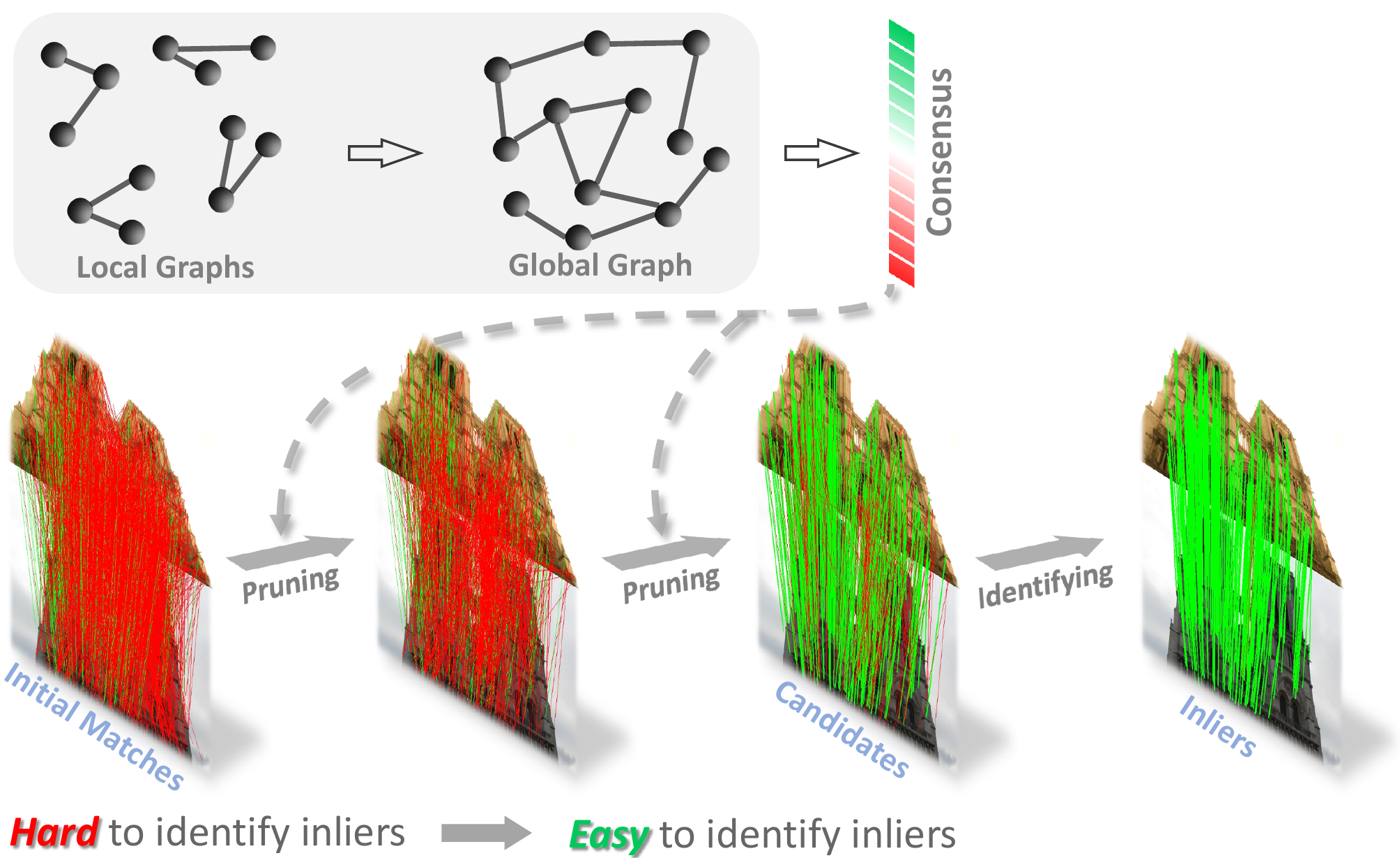}}
    \end{center}
 %   \vspace{-5pt}
   \caption{\textbf{Progressive correspondence pruning via local-to-global consensus learning.} Given initial correspondences (bottom-left image) with dominant {\color{red}outliers}, correctly identifying {\color{green}inliers} remains challenging. Instead of classifying correspondences in a one-shot fashion, we propose to gradually prune the correspondences to obtain a subset of reliable candidates based on correspondence consensus scores estimated from local-to-global graphs, encouraging accurate inlier identification.}
   \vspace{-8pt}
\label{fig:insight}
\end{figure}

Accurate pixel-wise correspondences act as a premise to tackle many important tasks in computer vision and robotics, such as Structure from Motion (SfM)~\cite{snavely2008modeling}, Simultaneous Location and Mapping (SLAM)~\cite{mur2015orb}, image stitching~\cite{brown2007automatic}, visual localization~\cite{philbin2010descriptor}, and virtual reality~\cite{szeliski1994image}. Unfortunately, feature correspondences established by off-the-shelf detector-descriptors ~\cite{lowe2004distinctive,mur2015orb,mishchuk2017working,detone2018superpoint} tend to be sensitive to challenging cross-image variations, such as rotations, scale changes, viewpoint changes, and illumination changes. 
Much recent research has therefore focused on correspondence selection~\cite{bian2017gms, ma2019locality, moo2018learning}, aiming to identify correct matches (inliers) while rejecting false ones (outliers).

\begin{figure*}[t]
    \begin{center}
        {\includegraphics[width=0.9\linewidth]{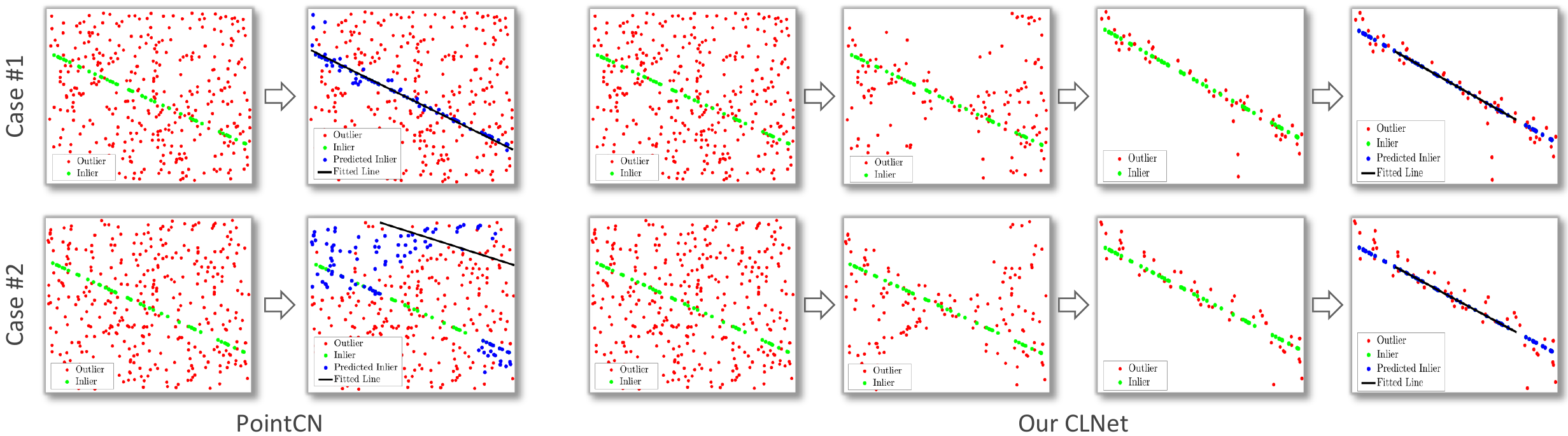}}
    \end{center}
    \vspace{-10pt}
   \caption{\textbf{Robustness against different distributions of outliers.} The {\color{green}inliers} are the same in both cases while the {\color{red}outliers} are randomly sampled. PointCN fails to find the correct line in the second case, showing its lack of robustness to the outlier distribution. By contrast, by gradually pruning the 2D points into reliable candidates for further line fitting, our method mitigates the effects of randomly-sampled outliers and consistently identifies the line.}
   %The introduced progressive pruning allows our method to be better generalized to different scenarios.}
   \vspace{-8pt}
%   The inliers (green dots) are sampled from the same line in Case 1 and Case 2, while outliers (red dots) are randomly sampled which leads to two different distributions in two cases. The line is well fitted by PointCN in Case 1, while the result is inferior in Case 2. By contrast, our method manages to fit the correct line in both cases, facing the challenge of different outlier distributions. }
\label{fig:line_visual}
\end{figure*}

In this context, deep learning has been utilized as a powerful solution~\cite{moo2018learning,zhao2019nm,zhang2019learning,sun2020acne}, typically casting correspondence selection as a per-match classification task, and adopting Multi-Layer Perceptrons (MLPs) to classify putative matches into inliers and outliers. %The optimization of such a binary classification problem is non-trivial, since
However, the resulting learning problem is significantly complicated by the fact that the initial matches are general extremely unbalanced, with around $90\%$ of outliers~\cite{zhao2019nm}(refer to the bottom-left image in Fig~\ref{fig:insight}), which are randomly distributed in real-world scenarios~\cite{zhao2019nm}. We therefore employ a toy line-fitting example shown in Fig.~\ref{fig:line_visual} to explain this issue, in which 100 inliers are identically sampled from the same line, while 900 outliers are randomly located. The standard PointCN~\cite{moo2018learning} baseline may fail to detect the same set of inliers depending on the outliers included in the data, because, given a finite training time, learning a distinctive feature embedding from arbitrarily located outliers is non-trivial.
%for networks.
%It requires the model to fit a line from given data points corrupted by randomly sampled outliers. Note that we adopt identical inliers and different outliers in two cases. We observe that the baseline method PointCN~\cite{moo2018learning} is not robust and fails in the second case.
%, which indicates that it lacks robustness on handling various outliers.

In this paper, motivated by the classical $L_\infty$ minimization method~\cite{sim2006removing}, we propose to \emph{progressively prune} the initial set of correspondences into a subset of candidates instead of classifying correspondences in a one-shot fashion. As the majority of outliers are expected to be filtered out after progressive pruning, this approach lets us identify reliable inliers among the candidates (bottom-right image in Fig.~\ref{fig:insight}). 
%However, the progressive process tends to be suboptimal without a reliable pruning method. 
This process, however, requires defining a pruning strategy. Leveraging the intuition that one cannot classify an isolated correspondence as inlier or outlier without context information, we therefore introduce a \textit{local-to-global consensus learning} framework, which \textit{explicitly} captures local and global correspondence context via \textit{dynamic graphs} to facilitate the correspondence pruning process. %Because the true inliers are not randomly distributed, they will receive larger support than the outliers, and thus can be identified despite being significantly outnumbered by the outliers. Such gradual denoising, however, cannot be achieved without exploiting contextual information, and we therefore introduce a \textit{local-to-global consensus learning} framework, which \textit{explicitly} captures local and global correspondence context to facilitate the correspondence pruning process. \MS{Pruning seems more appropriate to me than denoising. Denoising suggests that you modify the observations to remove the noise. You may want to consider replacing denoising with pruning throughout the whole paper.}

%Given the observed limitations of existing learning-based methods, we argue that properly \textit{denoising} correspondences is crucial for robust model estimation, \ie gradually tailor the noisy initial matches into reliable candidates (see Fig.~\ref{fig:insight}). Such an intuition is inspired by the classic algorithm RANSAC~\cite{fischler1981random}, whose core idea is to sample the most reliable subset with sufficient inliers iteratively. The introduced correspondence denoising could largely mitigate the effects of unbalanced initial matches and stably improves the robustness of model estimation, since inliers are expected to account for a larger proportion in the selected subset than the one in raw matches. Apparently, a distinctive pruning is a cornerstone of the correspondence denoising. As one cannot classify given an isolated correspondence without context information, we propose to operate the pruning by a \textit{local-to-global consensus learning} framework, which \textit{explicitly} captures local and global context for correspondences.

Specifically, we \textit{dynamically} construct a local graph for each input correspondence, whose nodes and edges represent the neighbors of the correspondence and their affinities in feature space, respectively. We then introduce an \textit{annular convolutional layer} to aggregate local features and produce a consensus score for each local graph. Guided by local consensus scores, we further merge multiple local graphs into a global one, from which we obtain a global consensus score via a spectral graph convolutional layer~\cite{kipf2016semi}. Together, the local and global consensus learning layers form a novel ``pruning'' block, which preserves potential inliers with higher consensus scores while filtering out outliers with lower scores. Correspondence pruning is then progressively achieved by stacking multiple pruning blocks. Such an architecture design encourages the refinement of local and global consensus learning at multiple scales. In contrast to previous works~\cite{moo2018learning,sun2020acne} that \textit{implicitly} model contextual information via feature normalization, our network \textit{explicitly} exploits context thanks to our local-to-global graphs.
% demonstrate its effectiveness to encourage better correspondence denoising. We exploit context information with novel local-to-global graphs for a totally different purpose.
Our contributions can be summarized as follows.

\vspace{-5pt}
 \begin{itemize}
 	\item 
	 We propose to progressively prune correspondences for better inlier identification, which alleviates the effects of unbalanced initial matches and random outlier distribution.
	 
	 \vspace{-5pt}
 	\item 
	We introduce a local-to-global consensus learning network for robust correspondence pruning, achieved by establishing dynamic graphs on-the-fly and estimating both local and global consensus scores to prune correspondences\footnote{Code is available at: \url{https://sailor-z.github.io/projects/CLNet}}.
	
	\vspace{-5pt}
 	\item 
	Our approach explicitly captures contextual information to identify inliers from outliers.
 \end{itemize}
We empirically demonstrate the effectiveness of our method on the tasks of robust line fitting, camera pose estimation and retrieval-based image localization. Our approach outperforms the state-of-the-art methods by a considerable margin.

\section{Related Work}

\noindent\textbf{Generation-verification framework.} The generation-verification framework has been widely used for robust model estimation, e.g., RANSAC~\cite{fischler1981random}, LO-RANSAC~\cite{chum2003locally}, PROSAC~\cite{chum2005matching},  USAC~\cite{raguram2012usac}, NG-RANSAC~\cite{brachmann2019neural}, \textit{etc}. It iteratively generates hypotheses and verifies the hypothesis confidence. Specifically, RANSAC~\cite{fischler1981random} randomly samples a minimal subset of data to estimate a parametric model, and then verifies its confidence by evaluating the consistency between the data and generated parametric model. NG-RANSAC~\cite{brachmann2019neural} proposes a two-stage approach which improves the sampling strategy of RANSAC by a pre-trained deep neural network. RANSAC and its variations have been proven sensitive to outliers in the literature~\cite{zhao2020image}, since the sampled subset is prone to including inevitable outliers in the case of consumed data extremely unbalanced with enormous outliers. They appear as powerful solutions of robust model estimation when the majority of outliers is removed in advance~\cite{zhao2020image, jin2020image} by correspondence pruning methods~\cite{moo2018learning, zhao2019nm, zhang2019learning}.

\noindent\textbf{Per-match classification.}
Inspired by the tremendous success of deep learning~\cite{he2016deep,ren2015faster,krizhevsky2017imagenet}, correspondence selection is now typically performed using deep networks. Due to the irregular and unordered characteristics of correspondences, 2D convolutions cannot be easily applied. PointCN~\cite{moo2018learning} proposes to treat the correspondence pruning as a per-match classification problem, using MLPs to predict the label (inlier or outlier) for each correspondence. Since then, per-match classification has become the de facto standard. NM-Net~\cite{zhao2019nm} expects to extract reliable local information for correspondences via a compatibility-specific mining, which relies on the known affine attributes. OANet~\cite{zhang2019learning} presents differentiable pooling and unpooling techniques to cluster input correspondences and upsample the clusters for a full size prediction (predicting labels for all input correspondences), respectively. An attentive context normalization is proposed in~\cite{sun2020acne}, which implicitly represents global context by the weighted feature normalization.
% While treated as a specific category, outliers are randomly distributed and make no contribution for high-level reconstruction or localization, which play a role of noise in practice. 
% Under the formulation of full-size per-match classification, the effect of outliers is inevitable, which limits the applications in the situation of outliers taking a tremendous part.
Although existing methods have shown satisfactory performance, they still suffer from the dominant outliers included in the putative correspondences. To address this issue, we suggest progressively pruning correspondences into a subset of candidates for easier inlier identification and more robust model estimation. We illustrate empirically that our approach effectively mitigates the effects of outliers in Sec.~\ref{sec:exper}.
% reformulating the per-match classification as a generation-verification framework, building upon local-to-global consensus learning. Only reliable candidates are sampled while unreliable matches are hierarchically removed, which explicitly alleviates the effect of outliers.

\noindent\textbf{Consensus in correspondences.} 
Correct matches are consistent in epipolar geometry or under the homography constraint~\cite{hartley2003multiple}, while mismatches are inconsistent because of their random distribution. The idea of correspondence consensus has therefore been studied, but mostly within hand-crafted methods. For instance, GTM~\cite{albarelli2012imposing} computes a game-theoretic matching based on a payoff function that utilizes affine information around keypoints to measure the consistency between a pair of correspondences. LPM~\cite{ma2019locality} assumes that local geometry around correct matches does not change freely; the geometry variation is represented by the consensus of $k$-nearest neighbors on keypoint coordinates. GMS~\cite{bian2017gms} proposes to indicate the consensus by the number of correspondences located in small regions. However, in~\cite{zhao2020image}, the hand-crafted methods 
are shown to be sensitive to specific nuisances such as rotation, translation, and viewpoint changes. Inspired by these hand-crafted efforts, we also exploit the notion of consensus, but propose to  \textit{learn} it using local-to-global graphs.

% \vspace{-5pt}
\section{Method}
\begin{figure}[t]
    \begin{center}
  %      {\includegraphics[width=1.0\linewidth]{figures/architecture.pdf}}
    %     \subfigure[Pipeline]
    % 	{ \includegraphics[width=0.53\linewidth]{figures/architecture.pdf}\label{fig:pip} }
    % 	\subfigure[Down-sampling block]
    % 	{ \includegraphics[width=0.4\linewidth]{figures/ds_block.pdf}\label{fig:ds} }
    \includegraphics[width=1.0\linewidth]{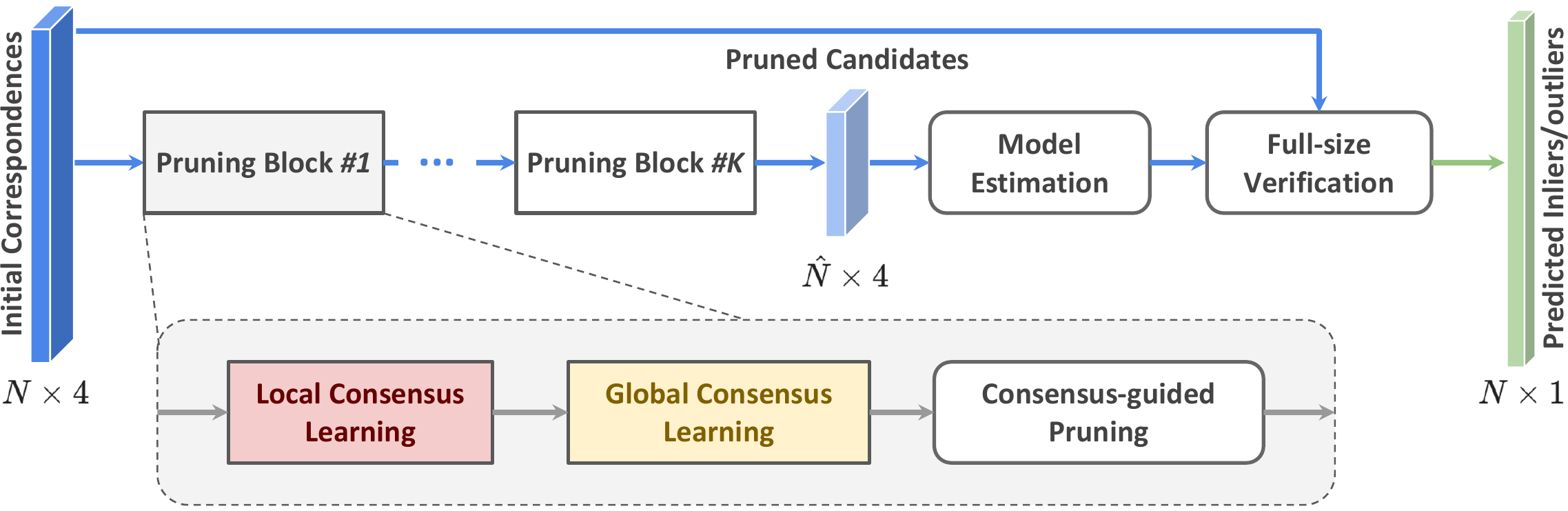}\label{fig:pip}
    \end{center}
    \vspace{-5pt}
   \caption{\textbf{Overall framework.} $N$ represents the number of matches and $4$ denotes the 4D locations of matched keypoints. We gradually prune the raw data into $\hat{N}$ candidates via $K$ pruning blocks guided by local-to-global consensus learning. A parametric model is then estimated, employing inliers identified among the $\hat{N}$ candidates. A full-size verification is further conducted based on the estimated model, yielding $N\times 1$ inlier/outlier predictions for the initial correspondences.}
    \vspace{-15pt}
\end{figure}

Let us introduce our local-to-global \textbf{C}onsensus \textbf{L}earning framework (\textbf{CLNet}) to tackle the challenge of dealing with massive amounts of outliers in a putative set of correspondences.
%(see Fig.~\ref{fig:pip}). 
The key innovation of our framework lies in progressively pruning the putative correspondences into more reliable candidates by exploiting their consensus scores. As illustrated in Fig.~\ref{fig:pip}, we achieve the progressive pruning via sequential ``pruning'' blocks that learn consensus using dynamic local-to-global graphs. Inliers are then identified from the pruned candidates and employed to estimate a parametric model. The parametric model is subsequently used to conduct a full-size verification~\cite{fischler1981random} for the complete set of putative correspondences. Below, we discuss our framework in detail. %In order to encourage a continuous latent space for more accurate consensus estimation, we introduce to use adaptive temperatures in the inlier/outlier classification losses as the training objective.\MS{I don't think this is necessary here.}

% Given a denoised subset of matches, the inliers can be easily identified, and a more accurate essential matrix can be recovered, where in turn, the essential matrix is employed to predict inliers and outliers out of all the initial correspondences. Such a ``generation-verification'' pipeline is inspired by classic and effective algorithm, RANSAC \cite{}.

% Fig.~\ref{fig:archi} illustrates the network architecture of our method. Initial correspondences are hierarchically down-sampled into a subset of candidates, which are employed to generate parametric hypothesis. A local-to-global consensus learning module is proposed, targeting to enlarge the response of inliers while down-sampling. The full size prediction is performed by verifying the generated hypothesis on initial data.  

% In the reminder of this section, we state the problem of correspondence selection and introduce the proposed method. The motivation of our hierarchically down-sampling framework and the details of our local-to-global consensus learning approach are described.

% In the reminder of this section, we first state the formulation of correspondence selection problem and then introduce our 

\subsection{Problem Formulation}
Given an image pair $(\mathbf{I}, \mathbf{I}{'})$, putative correspondences $\mC$ can be established via nearest neighbor matching between the descriptors of extracted keypoints. Let us denote the correspondences as $\mC=[{\vc_1}, \cdots, \vc_N] \in \mathbb{R}^{N\times 4}$.
% \begin{align}
%     \mC=[{\vc_1}, \cdots, \vc_N] \in \mathcal{R}^{N\times 4},
% \end{align}
% where 
$\vc_i=[x_i, y_i, x'_i, y'_i]$ indicates a correspondence between keypoint $(x_i, y_i)$ in image $\mathbf{I}$ and keypoint $(x'_i, y'_i)$ in image $\mathbf{I'}$.
Any off-the-shelf detectors and descriptors can be used for this task, either handcrafted methods~\cite{lowe2004distinctive, rublee2011orb} or learned ones~\cite{mishchuk2017working,detone2018superpoint}. In any event, the putative correspondences $\mC$ often contain a huge proportion of mismatches, and correspondence pruning thus aims to identify the correct matches (inliers) $\mC_p$ while rejecting the incorrect ones (outliers) $\mC_n$.
% where the inliers can be used for recovering the essential matrix $\hat{\mE}$. 
% Subsequently, an essential matrix can be recovered from the selected inliers $\mC_p$.

Existing learning-based methods~\cite{moo2018learning,sun2020acne} typically cast the correspondence pruning as an inlier/outlier classification problem, adopting permutation-invariant neural networks to predict inlier weights $\vw=\tanh(\text{ReLU}(\vo))\in[0,1)$ for all putative correspondences $\mC$, where $\vo$ is the output of network. %Note that only $\mC$ serves as input to the network.
$\vc_i\in\mC$ is then categorized as an outlier if its predicted weight $\evw_i=0$.
% A correspondence $\vc_i\in\mC$ will be categorized into outliers $\mC_n$ if its predicted weight $w_i=0$.
% Formally, we denote a correspondence as $\vc_i\in\mC$ and its predicted weight as $w_i\in[0,1)$. $\vc_i$ will be categorized into outliers $\mC_n$ if $w_i=0$.
% adopt permutation-invariant neural networks to predict a weight $w\in[0,1)$ for each putative correspondence, where $w=0$ indicates an outlier.
% , denoted as
% \begin{align}
%     \mC=[{\vc_1}, \cdots, \vc_N] \in \mathcal{R}^{N\times 4},
% \end{align}
% where $\vc_i=[x_i, y_i, x'_i, y'_i]$ indicates a correspondence between a keypoint $(x_i, y_i)$ in the image $\mathbf{I}$ and another keypoint $(x'_i, y'_i)$ in the paired image $\mathbf{I'}$.
% The keypoint coordinates are normalized by using camera intrinsics~\cite{moo2018learning}.
The predicted weights $\vw$ are not only utilized to identify inliers but also as auxiliary input for model estimation, \textit{e.g.}, to compute the essential matrix $\hat{\mE}$ for camera pose estimation~\cite{hartley2003multiple}. In other words, predicting accurate weights $\vw$ is at the core of learning-based correspondence pruning methods. This, however, is complicated by the dominant presence of outliers in $\mC$.
%Specifically, the initial matches are prone to being extremely unbalanced, with over 90\% of outliers~\cite{zhao2019nm} which are randomly distributed. 
Furthermore, the fact that existing methods~\cite{moo2018learning, sun2020acne} do not explicitly model the contextual information across $\mC$ makes the pruning process even more difficult.

\begin{figure*}[t]
    \begin{center}
%	\subfigure[]
%	{ \includegraphics[width=0.3\linewidth]{figures/annular.pdf}\label{fig:ds} }
%	\subfigure[]
	{ \includegraphics[width=0.9\linewidth]{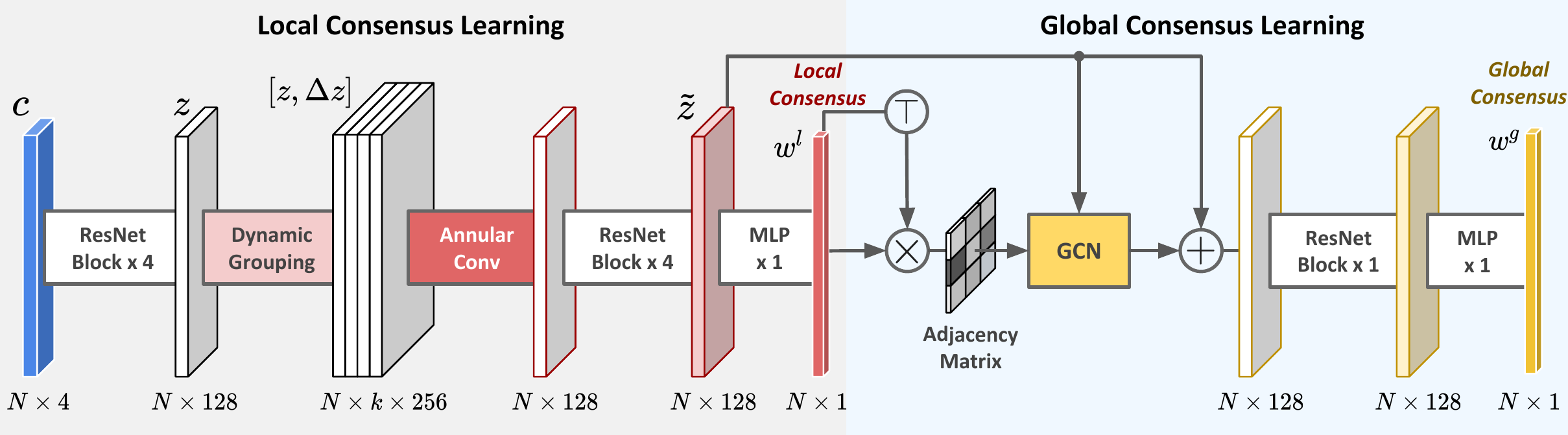}}
    \end{center}
    \vspace{-8pt}
   \caption{\textbf{Detailed architecture of the proposed pruning block.} A pruning block consists of local-to-global consensus learning layers. 
   Each ResNet block~\cite{he2016deep} contains two MLPs followed by Context Normalization~\cite{moo2018learning}, Batch Normalization~\cite{ioffe2015batch}, and ReLU.
%   Nine ResNet blocks \cite{} are utilized in each module, which results in the same network depth as~\cite{zhang2019learning}. 
%   Two MLPs followed by Context Normalization, Batch Normalization, and ReLU are employed in each ResNet layer. 
   Note that Attentive Context Normalization~\cite{sun2020acne} is not used in our method, because it requires additional supervision. We select a subset of candidates from input correspondences according to the estimated global consensus scores.}
   \vspace{-6pt}
\label{fig:archi}
\end{figure*}

In this paper, we advocate for progressively pruning $\mC$ into a subset of candidates $\hat{\mC}$, mitigating the effect of dominant presence of outliers. By predicting inlier weights $\hat{\vw}$ for the pruned subset $\hat{\mC}$ which is expected to be more reliable than $\mC$, inliers $\hat{\mC}_p$ can be obtained more easily. We then estimate a parametric model (essential matrix $\hat{\mE}$ as an example) from $\hat{\mC}_p$. In turn, this model is adopted to perform a full-size verification on the entire set $\mC$. Some inliers falsely rejected by pruning are expected to be recovered after the verification. Our solution therefore follows the ``generation-verification'' paradigm, as the classical and effective RANSAC algorithm~\cite{fischler1981random}.
% Specifically, training a network to perform full-size prediction directly \cite{} is sub-optimal, since the initial matches are extremely unbalanced with over 90\% outliers for each image pair.
% Moreover, it cannot be generalized well to deal with arbitrary outliers in real-world scenarios, as the outliers are randomly distributed which might not be ``seen'' from the training samples. 
% How to improve the robustness to random and dominant outliers is crucial to the final performance.

Formally, our approach can be expressed as
\vspace{-5pt}
\begin{align}
% \hat{\mC} = d_{\theta}(\mC),& ~~~
% \hat{\vw} = f_{\phi}(\hat{\mC}), \nonumber \\
& \hat{\mC} = f_{\phi}(\mC),~~~~(\hat{\vw}_p, \hat{\mC}_p) =f_{\psi}(\hat{\mC}) \nonumber \\
& \hat{\mE} = g(\hat{\vw}_p, \hat{\mC}_p),~~~~\vw = h(\hat{\mE}, \mC),
\label{eq:eq1}
\end{align}
\vspace{-15pt}

\noindent where 
% $d_\theta$ is the denoising module with learnable parameters $\theta$ and $f_\phi$ is the weight predicting module with learnable parameters $\phi$. 
$f_\phi$ and $f_\psi$ are deep neural networks with learnable parameters $\phi$ and $\psi$ that perform correspondence pruning and inlier identification, respectively; $g(\cdot,\cdot)$ denotes the parametric model estimation (generation) process; the optional $h(\cdot,\cdot)$ performs a full-size verification (prediction).% \MS{${\mC}_p$ does not appear in this formalism...}

\subsection{Local-to-Global Consensus Learning}
To design the network $f_{\phi}()$, we leverage the intuition that inliers should be consistent in both their local and global contexts, and thus propose to estimate consensus scores from local-to-global graphs. Correspondences with high scores are preserved whereas those with low scores are removed as outliers. %The consensus scores further serve as inlier weights; therefore we use the unified notation $\vw$ for these two concepts.
% Context information has been proven effective to determine inliers with geometrically consistent features.
% Existing methods \cite{} leveraging local context generally

As illustrated in Fig.~\ref{fig:pip}, our framework includes a sequence of ``pruning'' blocks, which progressively prune correspondences, \textit{i.e.}, $\mC\to\cdots\to\hat{\mC}$. Each block consists of local and global consensus layers, the detail of which is shown in Fig.~\ref{fig:archi}. In what follows, without loss of generality and for ease of notation, we denote the input to a ``pruning'' block as $\mC\in \mathbb{R}^{N\times 4}$ and its output as $\hat{\mC}\in {\mathbb{R}}^{\hat{N}\times 4}$, where $\hat{N}<N$. Note, however, that each block truly takes as input the output of the previous block. Let us now define the operations of a pruning block in more detail.
% In order to acquire $\mathcal{C}_k$ from $\mathcal{C}$, we propose to sample correspondences based on the consensus from local regions to a global region. The architectures of our consensus learning networks are shown in Fig.~\ref{fig:local} and Fig.~\ref{fig:global}. The local and global contexts are $explicitly$ captured by dynamic graphs. The details are introduced as follows.
\paragraph{Local consensus.}
% As discussed in~\cite{ma2014robust, zhao2019nm, ma2019locality}, the consensus in local regions is promising to determine inliers. However,~\cite{ma2014robust, ma2019locality} measure the consensus in Euclidean space, which is sensitive to specific noise~\cite{zhao2020image};~\cite{zhao2019nm} requires known affine information to estimate the consensus. Alternatively, we suggest predicting the consensus from local graphs which are $adaptively$ constructed in the embedded feature space.
We propose to leverage local context for each correspondence $\vc_i$ by building a $k$-nearest neighbor graph denoted as 
\begin{align}
    \gG_i^l = (\gV_i^l, \gE_i^l), ~~~1\le i\le N,
\end{align}
where $\gV_i^l=\{\vc_i^1,\cdots,\vc_i^k\}$ are $k$-nearest neighbors of $\vc_i$ in feature space, and $\gE_i^l$ indicates the set of directed edges that connect $\vc_i$ and its neighbors in $\gV_i^l$. Specifically, given $\vc_i\in\mC$, we extract a feature representation $\vz_i$ via a series of ResNet blocks~\cite{he2016deep}.
The $k$-nearest neighbors of $\vc_i$ are determined by ranking the Euclidean distances between $\vz_i$ and $\{\vz_j|1\le j\le N,j\ne i\}$.
Following~\cite{wang2019dynamic}, we describe the features between $\vc_i$ and each neighbor as
\begin{align}
    \ve_i^{j}=[\vz_i, \Delta\vz_i^{j}], ~~~1\le j \le k
\end{align}
where $[\cdot, \cdot]$ represents the concatenation; $\Delta\vz_i^{j}=\vz_i-\vz_i^j$ is the residual feature of $\vc_i$ and the $j$-th neighbor $\vc_i^j$.

% We denote that $\mathbf{f}_i$ is the embedded feature of $\mathbf{c}_i$ estimated as 
% \begin{equation}\label{eq:LRF2}
% \mathbf{f}_i=f(\mathbf{c}_i;\Omega),
% \end{equation}
% where $\Omega$ is the set of learned parameters. The local graph $\mathcal{G}_i^{l}=(\mathcal{V}_i^{l}, \mathcal{E}_i^{l})$ around $\mathbf{c}_i$ is generated, building upon the similarity of $\mathbf{f}$. $\mathcal{V}_i^{l}$ and $\mathcal{E}_i^{l}$ represent the sets of nodes and edges in $\mathcal{G}_i^{l}$, respectively. Specifically, $\mathcal{V}_i^{l} = \{c_1, ..., c_k\}$ is grouped by $k$-nearest neighbor ($knn$) search according to the distance $d_{ij}=\left\|\mathbf{f}_i-\mathbf{f}_j\right\|_{2}$ where $\left\|\cdot\right\|_{2}$ indicates $L_2$ norm. $\mathbf{e}_{ij}^{l}\in\mathcal{E}_i^{l}$ describes the relationship between $\mathbf{c}_i$ and $\mathbf{c}_j$, computed as
% \begin{equation}\label{eq:LRF3}
% \mathbf{e}_{ij}^{l}=cat(\mathbf{f}_i, \mathbf{f}_i - \mathbf{f}_j),
% \end{equation}
% where $cat$ represents concatenation as suggested in~\cite{wang2019dynamic}.
Our goal then is to compute a local consensus score from the local graph $\gG_i^l$ for $\vc_i$. Intuitively, we split such a process into two steps: 1) Aggregating the features $\{\ve_i^j|1\le j\le k\}\to\tilde{\vz}_i$ by passing messages along graph edges $\gE_i^l$, and 2) predicting a consensus score from $\tilde{\vz}_i$ via MLPs. A na\"ive way for feature aggregation consists of using MLPs followed by pooling layers~\cite{wang2019dynamic}. However, this operation may discard the structural information in the graphs, \textit{i.e.}, the fine-grained relations among graph nodes, because the $1\times1$ kernels of MLPs extract features from neighbors separately. %the $k$-nearest neighbors in $\gV_i^l$ are actually sorted by their affinities and should be treated differently.
\begin{figure}[t]
    \begin{center}
        {\includegraphics[width=0.7\linewidth]{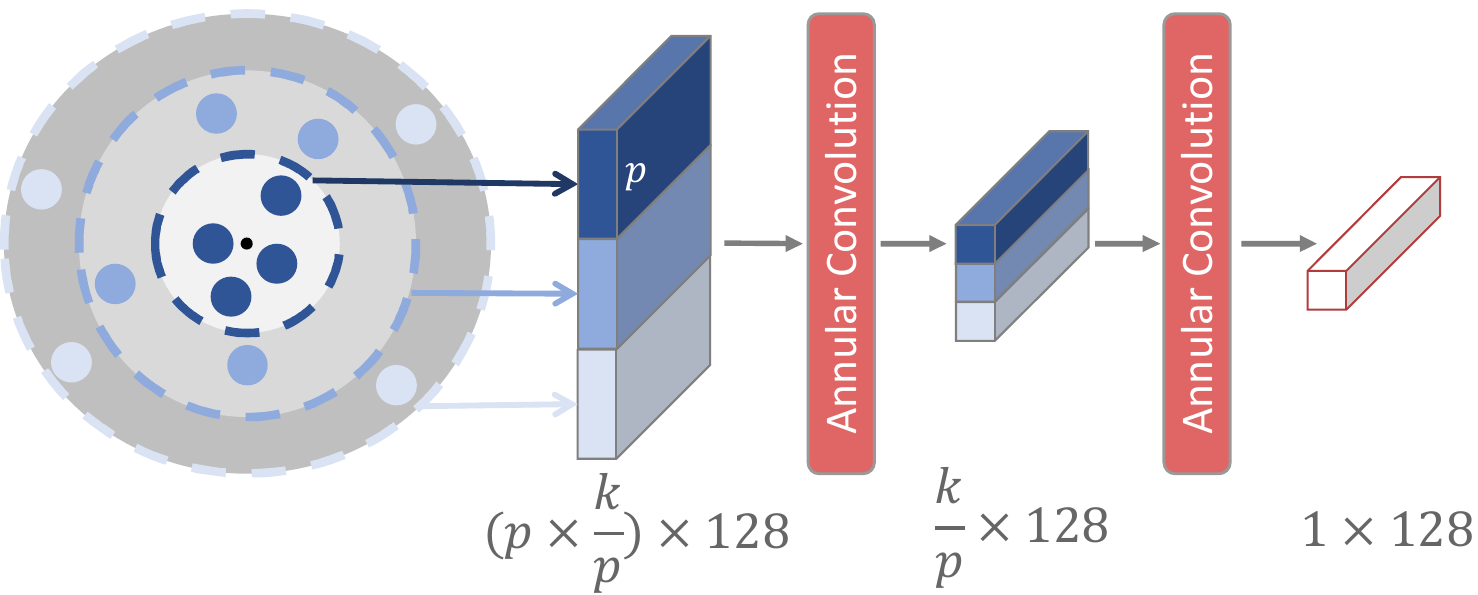}}
    \end{center}
    \vspace{-10pt}
   \caption{\textbf{Illustration of the proposed annular convolution.} The nodes (colored dots) in a local graph are grouped into annuli based on their affinities to the anchor. The features in each annulus are aggregated by a convolution kernel.}
\label{fig:annular}
\vspace{-6pt}
\end{figure}
To make the most of the graph knowledge, we therefore introduce an \textit{annular} convolutional layer which considers both the affinities of neighbors to the anchor and the relative relationships among neighbors in each annulus. Specifically, as shown in Fig.~\ref{fig:annular}, nodes are sorted based on the affinities to the anchor and then assigned into $k / p$ annuli, where $p$ denotes the number of nodes in each annulus and $k$ is expected to be divisible by $p$. We aggregate the features in each annulus via $1 \times p$ convolution kernels as
\begin{align}
\label{eq:LRF4}
     \tilde{\ve}_i^{t} = \sum\mathbf{W}\ve_i^{j} + \mathbf{b}, ~~~(t-1)p\le j \le tp
\end{align}
where $\tilde{\ve}_i^{t}$ denotes the aggregated feature of the $t$-th annulus; $\mathbf{W}$ and $\mathbf{b}$ are learnable parameters. $\{\tilde{\ve}_i^{t}|1\le t\le k/p\}$ are further integrated by another annular convolution, \textit{i.e.}, the second one in Fig.~\ref{fig:annular}, with its own set of trainable parameters. These operations are followed by ResNet blocks to extract a feature vector $\tilde{\vz}_i$, which we further transform into a local consensus score $\evw_i^l$. $\evw_i^l$ reflects the consistency of $\vc_i$ in the local receptive field. In other words, $\evw_i^l$ roughly measures the inlier weight of $\vc_i$ when only considering its local context.

% And so on, we can finish feature aggregation by using $\sqrt[p]{k}$ annulus convolution layers sequentially. 
% We denote the aggregated feature as $\hat{z}_i$, showing the same dimensions as $z_i$.

% Further, the core issue is estimating the consensus from $\mathcal{G}_i^{l}$. Intuitively, the consensus is related to the neighbor distribution in the feature space. Consequently, we present an $annular$ convolution to capture fine-grained patterns with different scales. As shown in the dash box in Fig.~\ref{fig:local}, we separate $\mathcal{V}_i^{l}$ into some nonoverlapping subsets based on $d_{ij}$. The feature variation in each subset is then captured as 
% \begin{equation}\label{eq:LRF4}
% \mathbf{f}=\sum_{j=1}^{a}\mathbf{W}^{l}\mathbf{e}_{ij}^{l} + \mathbf{b}^{l},
% \end{equation}
% where $a$ is the size of each subset, $\mathbf{W}^{l}$ and $\mathbf{b}^{l}$ are learned parameters. The separation and aggregation are alternatively performed, which generates a general consensus description $\mathbf{s}_i^{l}$ of $\mathcal{G}_i^{l}$ from multiple $annular$ regions.
\vspace{-6pt}
\paragraph{Global consensus.}
To encode global contextual information, we connect the local graphs into a global one. The global graph is denoted as
\begin{align}
    \gG^g = (\gV^g, \gE^g),
\end{align}
where nodes $\gV^g$ are represented by $\mathbf{C}$ with local aggregated features $\{\tilde{\vz}_1,\cdots,\tilde{\vz}_N\}$, and edges $\gE^g$ connect every two correspondences $(\mathbf{c}_i, \mathbf{c}_j), 1 \le i, j \le N, i \ne j$.  We encode the affinity of $(\mathbf{c}_i, \mathbf{c}_j)$ using the local consensus scores as
%As inliers show promising compatibility~\cite{zhao2019nm} 
%and , an edge $\ve^g_{ij} \in \gE^g$ is computed by 
\begin{equation}
e^g_{ij} = \evw^{l}_i\cdot \evw^{l}_j, 1\le i,j \le N.
\end{equation}
The product of $(\evw^{l}_i, \evw^{l}_j)$ enables $e^g_{ij}$ to indicate the compatibility of $(\mathbf{c}_i, \mathbf{c}_j)$.
An adjacency matrix $\mathbf{A}\in\mathbb{R}^{N \times N}$ is then built with $\mathbf{A}_{ij}=e^g_{ij}$ to explicitly describe the global context. Specifically, we exploit $\mathbf{A}$ to compute a global embedding
\begin{equation}
\vf^{g} = \mathbf{L}[\tilde{\vz}_1, \cdots, \tilde{\vz}_N]\mathbf{W}^{g},
\end{equation}
\begin{equation}
\mathbf{L}=\widetilde{\mathbf{D}}^{-\frac{1}{2}}\widetilde{\mathbf{A}}\widetilde{\mathbf{D}}^{-\frac{1}{2}}, 
\end{equation}
%\MS{I don't like the name ``out" too much.}
where $\{\tilde{\vz}\}$ are the aggregated features obtained from local consensus learning; $\mathbf{W}^{g}$ is a learnable matrix; $\mathbf{L}$ is the graph Laplacian~\cite{kipf2016semi}; $\widetilde{\mathbf{A}}=\mathbf{A} + \mathbf{I}_N$ for numerical stability; $\widetilde{\mathbf{D}}\in\mathbb{R}^{N\times N}$ is the diagonal degree matrix of $\widetilde{\mathbf{A}}$. 
In short, $\mathbf{L}$ modulates $\{\tilde{\vz}\}$ into the spectral domain, considering the isolated local embeddings in a joint manner, and 
the feature filter $\mathbf{W}^{g}$ in the spectral domain enables the propagated features to reflect the consensus from the global graph Laplacian.
Similar to local consensus learning, global consensus scores $\vw^g$ are estimated by encoding the aggregated features via a ResNet block followed by MLPs.
\vspace{-10pt}
\paragraph{Consensus-guided pruning.}
Since the global consensus scores jointly consider both global and local context, we prune the putative correspondences based on their global consensus scores $\vw^g$. Specifically, the elements in $\mC$ are sorted by a descending value of $\vw^g$. We preserve the top-$\hat{N}$ correspondences and discard the remaining ones as outliers. The back propagation can be achieved by only keeping the gradients of top-$\hat{N}$ correspondences. Furthermore, inspired by~\cite{zhang2019learning}, we take local and global consensus scores as additional input to the next pruning block.
A ResNet block and MLPs are used after the last pruning block to predict inlier weights $\hat{\vw}$ for the pruned candidates.

\vspace{-8pt}
\paragraph{Training objectives.}
Learning-based correspondence pruning methods~\cite{moo2018learning, sun2020acne} generally combine an inlier/outlier classification loss and a regression loss as the training objective. For camera pose estimation, on widely-used benchmarks~\cite{moo2018learning,zhang2019learning,jin2020image}, the ground-truth labels $\vs$ of $\mC$ are assigned using the epipolar distances with an ad-hoc threshold $d_\text{thr}$, empirically set to $1e$-$4$~\cite{zhang2019learning}. 

Although training with a conventional binary cross-entropy loss has achieved satisfactory performance~\cite{moo2018learning,zhao2019nm,zhang2019learning}, we argue that inevitable label ambiguity exists, especially for the correspondences whose epipolar distances are close to $d_\text{thr}$. Typically, the confidence of $\vc_i$ should be negatively correlated with the corresponding epipolar distance $d_i$, 
\textit{i.e.} $d_i \to 0$ for an inlier. 
% \textit{e.g.} a most confident inlier has an extreme epipolar distance $d_i=0$.
To reflect this intuition, we introduce an adaptive temperature for putative inliers ($d_i < d_\text{thr}$) to alleviate the effect of label ambiguity, computed as
\begin{equation}
\tau_i=\exp({-\frac{\|d_i - d_\text{thr}\|_1}{d_\text{thr}}}).
\label{eq:tem}
\end{equation}
For outliers $\vc_i$ with $d_i >= d_\text{thr}$, we set $\tau_i=1$.% Note that erroneous label assignment cannot be eliminated due to the inherent ambiguity of the epipolar constraint~\cite{hartley2003multiple}. 

The overall training objective is then denoted as
\begin{align}
    \mathcal{L} = \mathcal{L}_\text{cls} + \lambda \mathcal{L}_\text{reg}({\hat{\mathbf{E}}, \mathbf{E}}),
    \label{eq:loss}
\end{align}
where $\mathcal{L}_\text{cls}$ is a binary classification loss with our proposed adaptive temperature, $\mathcal{L}_\text{reg}$ represents a geometric loss~\cite{zhang2019learning} on estimated parametric model $\hat{\mathbf{E}}$, and $\lambda$ is a weighting factor. $\mathcal{L}_\text{cls}$ is formulated as
\begin{align}
    \mathcal{L}_\text{cls} = \sum_{j=1}^{K}\Big(\ell_\text{bce}&(\sigmoid(\mathbf{\ttau}_j\odot\mathbf{o}_{j}^{l}), \mathbf{y}_{j}) + \ell_\text{bce}(\sigmoid(\mathbf{\ttau}_j\odot\mathbf{o}_{j}^{g}), \mathbf{y}_{j})\Big) \nonumber\\
    &+\ell_\text{bce}(\sigmoid(\hat{\mathbf{\ttau}}\odot\hat{\mathbf{o}}), \mathbf{y}),
\end{align}
where $\vo_{j}^{l}$, $\vo_{j}^{g}$ are the outputs of local and global consensus learning layers in $j$-th pruning block, respectively; $\hat{\mathbf{o}}$ is the output of the last MLP in CLNet; %(with $\vw=\tanh(\text{ReLU}(\vo))$); 
$\bf{\ttau}$ is a vector of temperatures estimated by Eq.~\ref{eq:tem}; $\sigmoid$ represents the sigmoid function; $\odot$ indicates the Hadamard product; $\mathbf{y}$ denotes the set of binary ground-truth labels; $\ell_\text{bce}$ indicates a binary cross-entropy loss; $K$ is the number of pruning blocks. 
As a result, an inlier $c_i$ with a smaller $d_i$ would be more confident to enforce larger regularization on the model optimization via a smaller temperature.% Note that all the local consensus scores $\{\vw^l\}$ and global consensus scores $\{\vw^g\}$ yield by multiple down-sampling blocks, as well as the final inlier weights $\hat{\vw}$ of the denoised candidates, are used together in the training objective to form the inlier/outlier classification loss. 

% The learned consensus score is weighted by the temperature and supervised as 
% \begin{equation}\label{eq:LRF11}
% \mathcal{L}_c = -\frac{1}{N}\sum_{i=1}^{N}y_i^{gt}log(S(t_i\cdot o_i)) + (1-y_i^{gt})log(1 - S(t_i\cdot o_i))
% \end{equation}
% For $c_i$ with a large $\Delta d_i$, the label assignment is convincing. In this context, a small $t_i$ forces networks to learn a distinctive $o_i$, i.e.,  $o_i \to +\infty$ for inliers; $o_i \to -\infty$ for outliers. By contrast, in the case of $\Delta \to 0$, the confidence of label assignment decreases. The large temperatures ease the learning procedure, which tightens the distribution of $o_i$.  
%\section{Relationship with Previous Works}
%\noindent\textbf{NG-RANSAC. }
%\noindent\textbf{NM-Net. }
%\noindent\textbf{OANet. }
%\noindent\textbf{ACNe. }

\section{Experiments}
\label{sec:exper}
We conduct experiments on four datasets, covering the tasks of robust line fitting (Section~\ref{sec:line}), camera pose estimation (Section~\ref{sec:matching}), and retrieval-based image localization (Section~\ref{sec:localization}). In Section~\ref{sec:analysis}, we provide a comprehensive analysis of our method to demonstrate the effectiveness of its components. 

\subsection{Implementation Details}
% \noindent\textbf{Implementation details.}
% In the following experiments, we employ an iterative design, as introduced in~\cite{zhang2019learning}. 
In our experiments, we use two sequential pruning blocks, pruning the $N$ putative correspondences into $N/4$ candidates, \textit{i.e.}, pruning by half in each block.
% The down-sampling procedure is iteratively performed twice, which down-samples $N$ correspondences into $N/4$ candidates. 
We set the number of nearest neighbors used to establish the local graphs to $k=9$ and $k=6$ for the two blocks, respectively. 
%Each annuli contains 3 nodes, \textit{i.e.} 
% In the local consensus learning network, we set $k=9$ during the first down-sampling, $k=6$ for the second one, and $a=3$ in Eq.~\ref{eq:LRF4}. 
We use $p=3$ for annular convolutions and $\lambda=0.5$ in Eq.~(\ref{eq:loss}). $h(\cdot)$ in Eq.~(\ref{eq:eq1}) is implemented by the epipolar constraint~\cite{hartley2003multiple} for camera pose estimation and image localization. For training, we use  the Adam~\cite{kingma2014adam} optimizer with a batch size of 32 and a constant learning rate of $10^{-3}$. 
%See Appendix \ref{} for the detailed network architecture.

\subsection{Robust Line Fitting}
\label{sec:line}
As a first experiment, to evaluate robustness to different outlier distributions, we focus on the task of robust line fitting~\cite{sun2020acne}. 
Note that this task differs from the correspondence selection problem, showing the ability of our method to address general inlier/outlier classification tasks. In this case, the input data to our network consists of 2D points $[x,y]$ instead of 4D correspondences $[x, y, x^{'}, y^{'}]$. Specifically, we create synthetic data by considering a 2D line $ax+by+c=0$ with parameters $(a,b,c)$  randomly sampled in $[0, 1]$. We generate inliers by randomly sampling $x \in [-5, 5]$ and estimating $y$ using the line equation. 
Outliers are obtained via uniformly randomly sampling $(x,y)$ in $[-5,5]$. For each line, we generate
$N=1000$ points, from which we seek to identify the inliers. We then compute $(a,b,c)$ using the least-square solution of~\cite{lawson1995solving}. $6000$, $2000$, and $2000$ data are used for training, validation, and testing, respectively. In addition to our method, we retrained PointCN~\cite{moo2018learning}, OANet~\cite{zhang2019learning}, and PointACN~\cite{sun2020acne} using the official implementations released by the authors. 

\begin{figure}[t]
    \begin{center}
        {\includegraphics[width=0.65\linewidth]{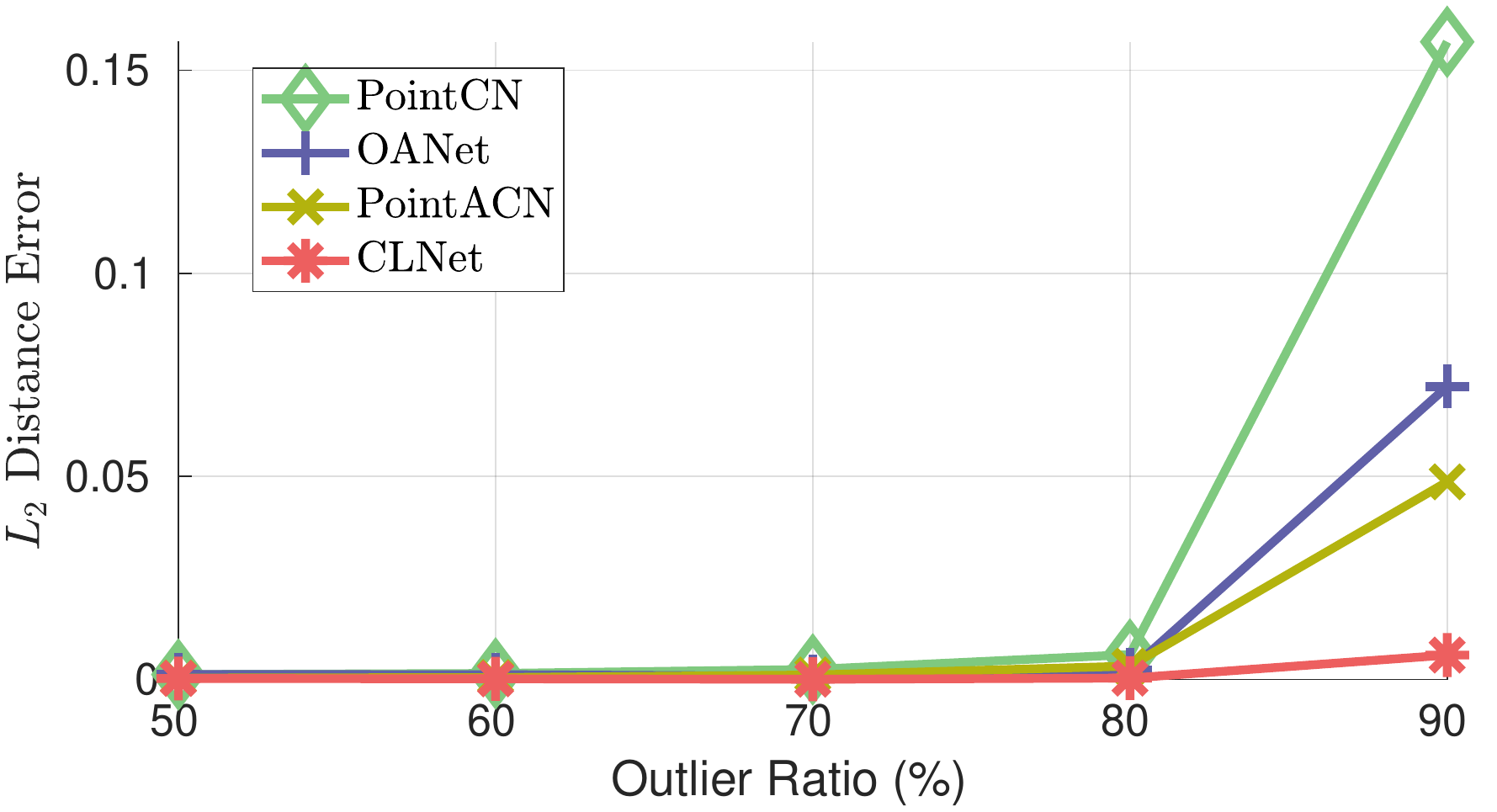}}
    \end{center}
    \vspace{-5pt}
   \caption{\textbf{Robust line fitting performance}. We report the $L_2$ distance between the predicted line parameters and the ground-truth ones for five different outlier ratios, varying from $50\%$ to $90\%$.}
   \vspace{-6pt}
\label{fig:line}
\end{figure}

Fig.~\ref{fig:line} summarizes the results of our line-fitting experiment for an outlier ratio ranging from $50\%$ to $90\%$. We report the $L_2$ distance between the ground-truth $(a, b, c)$ and the predicted ones. While all methods perform well for relatively low outlier ratios, ours significantly outperforms the competitors for high ones, \textit{i.e.} $90\%$ outliers. This evidences the benefits of pruning the input data instead directly trying to classify each input sample.

\subsection{Camera Pose Estimation}
\label{sec:matching}

For camera pose estimation, we exploit the outdoor YFCC100M~\cite{thomee2016yfcc100m} and indoor SUN3D~\cite{xiao2013sun3d} datasets, following the settings in~\cite{zhang2019learning, sarlin2020superglue} (more results on \textit{ScanNet}~\cite{dai2017scannet} and the benchmark of~\cite{zhang2019learning} can be found in the appendix). Initial matches are generated by nearest neighbor matching with SIFT~\cite{lowe2004distinctive}, unless otherwise specified.
%by default. %As suggested in ~\cite{moo2018learning, zhang2019learning,sun2020acne}, a weighted 8-point algorithm is used to compute the essential matrix ($\hat{\mE}$) from the denoised candidates $\hat{\mC}$, which is pivotal to recover camera poses. 
The keypoint coordinates in $\mC$ are normalized using the camera intrinsics, as in~\cite{moo2018learning}. %Epipolar distances of $\mC$ are estimated under the constraint of $\hat{\mathbf{E}}$ and then compared with a threshold ($d_\text{thr}=1e$-$4$ by default) to perform a full-size verification. 
Following~\cite{sarlin2020superglue}, %\MS{Have other papers also reported the same metrics?} 
we report the AUCs of the pose error at different thresholds ($5^{\circ}, 10^{\circ}, 20^{\circ}$). 
\begin{table}[t]
    \centering
    \footnotesize
    % \begin{tabular}{c}
    \subtable{
        \begin{tabular}{l|c|c c c}
        \hline
        \multirow{2}*{Method} & \multirow{2}*{Desc} & \multicolumn{3}{c}{YFCC100M~\cite{thomee2016yfcc100m} (outdoor) (\%)}  \\
        \cline{3-5}
        & & AUC@$5^{\circ}$ & AUC@$10^{\circ}$ & AUC@$20^{\circ}$ \\
        \hline
        RANSAC~\cite{fischler1981random} & \cmark & 14.33 & 27.08 & 42.27 \\
        MAGSAC~\cite{barath2019magsac} & \cmark & 17.01 & 29.49 & 44.03 \\
        LPM~\cite{ma2019locality} & \xmark & 10.22 & 20.65 & 33.96 \\
        GMS~\cite{bian2017gms} & \xmark & 19.05 & 32.35 & 46.79 \\
        CODE~\cite{lin2017code} & \xmark & 16.99 & 30.23 & 43.85 \\
        PointCN~\cite{moo2018learning} & \xmark & 26.53 & 43.93 & 61.01 \\
        NM-Net~\cite{zhao2019nm} & \xmark & 28.56 & 46.53 & 63.55  \\
        NG-RANSAC~\cite{brachmann2019neural} & \cmark & 27.17 & 43.60 & 59.63\\
        OANet~\cite{zhang2019learning} & \xmark & 28.76 & 48.42 & 66.18 \\
        PointACN~\cite{sun2020acne} & \xmark & 28.81 & 48.02 & 65.39 \\
        SuperGlue~\cite{sarlin2020superglue} & \cmark & 30.49 & 51.29 & 69.72 \\
        \hline
        Our CLNet & \xmark & \bf{32.79} & \bf{52.70} & \bf{69.76} \\
        \hline
        \end{tabular}
    }
    \subtable{
        \begin{tabular}{l|c c c}
        \hline
        \multirow{2}*{Method} & \multicolumn{3}{c}{SUN3D~\cite{xiao2013sun3d} (indoor) (\%)}  \\
        \cline{2-4}
        & AUC@$5^{\circ}$ & AUC@$10^{\circ}$ & AUC@$20^{\circ}$ \\
        \hline
        RANSAC~\cite{fischler1981random} & 3.93 & 10.28 & 21.04 \\
        MAGSAC~\cite{barath2019magsac} & 3.94 & 10.33 & 21.25 \\
      %  Ratio Test~\cite{lowe2004distinctive} & 4.51 & 11.62 & 23.02 \\
        LPM~\cite{ma2019locality} & 3.31 & 8.56 & 17.73 \\
        GMS~\cite{bian2017gms} & 4.36 & 11.08 & 21.68 \\
        CODE~\cite{lin2017code} & 3.52 & 8.91 & 18.32 \\
        PointCN~\cite{moo2018learning} & 5.86 & 14.40 & 27.12 \\
        NM-Net~\cite{zhao2019nm} & 6.45 & 16.44 & 31.16 \\
        NG-RANSAC~\cite{brachmann2019neural} & 6.65 & 16.46 & 30.86 \\
        OANet~\cite{zhang2019learning} & 6.83 & 17.10 & 32.28 \\
        PointACN~\cite{sun2020acne} & 7.10 & 17.92 & 33.56 \\
        \hline
        Our CLNet & \bf{7.78} & \bf{19.07} & \bf{35.25}  \\
        \hline
        \end{tabular}
    }
    \caption{\textbf{Pose estimation on YFCC100M~\cite{thomee2016yfcc100m} and SUN3D~\cite{xiao2013sun3d}.} ``Desc" indicates whether descriptors are required as input. For~\cite{fischler1981random, barath2019magsac, brachmann2019neural}, the ratio test of~\cite{lowe2004distinctive} is employed to prune the correspondences.}
    %AUCs of the pose error are reported.}
\label{tab:ess}
%\vspace{-5pt}
\end{table}

\begin{table}[t]
    \centering
    \small
    \begin{center}
        \begin{tabular}{l|c c c}
        \hline
        Method & AUC@$5^{\circ}$ & AUC@$10^{\circ}$ & AUC@$20^{\circ}$ \\
        \hline
        PointCN~\cite{moo2018learning} & 12.38 & 28.15 & 48.04 \\
        NM-Net~\cite{zhao2019nm} & 12.59 & 30.62 & 52.07  \\
        OANet~\cite{zhang2019learning} & 16.86 & 36.74 & 57.40 \\
        PointACN~\cite{sun2020acne} & 18.64 & 38.76 & 59.56 \\
        \hline
        Our CLNet & \bf{26.19} & \bf{46.33} & \bf{65.48} \\
        \hline
        \end{tabular}
    \end{center}
    \caption{\textbf{Pose estimation without a robust estimator on YFCC100M.} For these results, we used the weighted 8-point algorithm~\cite{moo2018learning} instead of a robust estimator, e.g., RANSAC, to estimate the essential matrices.}
\label{tab:dlt}
\end{table}

\begin{table}[t]
    \scriptsize
    \begin{center}
        \begin{tabular}{p{1.4cm}|p{1.69cm}<{\centering}|p{0.7cm}<{\centering}p{1.65cm}<{\centering}p{0.6cm}<{\centering}}
        \hline
         \multirow{2}*{Method} & {SUN3D~\cite{xiao2013sun3d} (\%)} & \multicolumn{3}{c}{YFCC100M~\cite{thomee2016yfcc100m} (\%)} \\
        % \hline
        \cline{2-5}
         & SIFT~\cite{lowe2004distinctive} & ORB~\cite{rublee2011orb} & DoG-Hard~\cite{mishchuk2017working} & SP~\cite{detone2018superpoint}\\
        \hline
        PointCN~\cite{moo2018learning} & 6.00 & 5.70 & 31.89 & 18.12 \\
        NM-Net~\cite{zhao2019nm}  & 5.78 & 5.26 & 32.70 & 17.43 \\
        OANet++~\cite{zhang2019learning}  & 5.60 & 6.35 & 32.53 & 17.75 \\
        PointACN~\cite{zhang2019learning}  & 5.55 & 5.56 & 32.02 & 17.62 \\
        \hline
        Our CLNet & \bf{6.25} & \bf{7.68} & \bf{35.10} & \bf{18.82} \\
        \hline
        \end{tabular}
    \end{center}
    \caption{\textbf{Generalization ability.} All models were trained on YFCC100M~\cite{thomee2016yfcc100m} with SIFT~\cite{lowe2004distinctive}, and tested on SUN3D~\cite{xiao2013sun3d} with SIFT, and on YFCC100M with ORB~\cite{rublee2011orb}, DoG-HardNet~\cite{mishchuk2017working}, or SuperPoint(SP)~\cite{detone2018superpoint}. AUC@$5^{\circ}$(\%) is reported.}
    \label{tab:gengelize}
    \vspace{-5pt}
\end{table}

Table~\ref{tab:ess} provides the quantitative results on YFCC100M and SUN3D. %\MS{Do they not report the numbers in their paper? Can't we directly use these numbers?}. 
For RANSAC~\cite{fischler1981random}, MAGSAC~\cite{barath2019magsac}, and NGRANSAC~\cite{brachmann2019neural}, we cleaned the initial correspondences using the ratio test of~\cite{lowe2004distinctive} with a threshold of $0.9$, because we observed the results without ratio test to be significantly worse. For the other methods, following~\cite{sarlin2020superglue}, we employ RANSAC as a robust estimator when estimating the essential matrices.
% We implement a $vanilla$ version of CLNet, which employs a PointCN-liked network to guide the down-sampling without consensus learning. 
%Note that SuperGlue~\cite{sarlin2020superglue} is not compared since it targets at another task, \textit{i.e.} predicting high-quality \textit{initial} matches rather than correspondence selection. 
Our CLNet delivers the best AUCs on both two datasets, even outperforms the most recent SuperGlue which requires descriptors as input (please refer to the appendix for more comparison with SuperGlue). 

We further consider the case of estimating the essential matrices without a robust estimator, \textit{i.e.}, without RANSAC. In this case, we use a weighted 8-point algorithm, as suggested in ~\cite{moo2018learning, zhang2019learning,sun2020acne}. As shown in Table~\ref{tab:dlt}, our CLNet achieves remarkably superior results, showing that our iterative pruning scheme makes our approach effective even without requiring an additional RANSAC step. 
% SUN3D dataset is much more challenging because the repeatable and low texture results in considerable ambiguity of the ground-truth label assignment.
%The ground-truth label assignments of SUN3D dataset are considerably ambiguous due to the repeatable and low texture~\cite{moo2018learning}, making RANSAC an important role for data post-processing~\cite{jin2020image}. Our CLNet-RANSAC combination outperforms all state-of-the-arts and CLNet alone achieves competitive performance on SUN3D. 
%Note that previous learning-based methods, \textit{e.g.} PointACN as well as OANet++, cannot be well compatible with RANSAC, showing inferior final performance, which would limit their applications in real-world scenarios. 
% By contrast, our method can be 
% The CLNet-RANSAC combination results in consistently positive impacts on SUN3D, while the competitors are empirically not compatible with RANSAC, \textit{e.g.} PointACN-RANSAC performs worse than PointACN. 
% CLNet is comparable with PointACN without RANSAC on SUN3D, since the repeatable and low texture causes considerable ambiguity of ground-truth label assignment.
% Compared with the $vanilla$ version, the guidance of consensus learning leads to a $7.72\%$ increase of mAP5 on YFCC100M, which shows the superiority of consensus guided framework.

To evaluate generalization ability, we test all learning-based methods on SUN3D with SIFT, and on YFCC100M with ORB~\cite{rublee2011orb}, DoG-HardNet~\cite{mishchuk2017working}, or SuperPoint~\cite{detone2018superpoint}, employing the models trained on YFCC100M with SIFT. Our choices of ORB, DoG-HardNet and SuperPoint were motivated by their popularity in SLAM~\cite{mur2015orb}, demonstrated robustness~\cite{jin2020image}, and joint detector/descriptor learning ability, respectively. As shown in Table~\ref{tab:gengelize}, CLNet achieves the best performance in all settings, which shows the robustness of our method to different datasets and detector-descriptor combinations.

\subsection{Retrieval-based Image Localization}
\label{sec:localization}

%Correspondences can be used for camera pose estimation~\cite{sattler2016large, taira2018inloc, brachmann2020visual} in image localization tasks, given the paired images that are identified first by image retrieval methods, \textit{e.g.}, NetVLAD~\cite{arandjelovic2016netvlad}.
Accurate retrieval is the premise of image localization~\cite{sattler2016large, taira2018inloc, brachmann2020visual}, providing initial locations of query images by identifying nearby reference images with geographical tags~\cite{ge2020self}.
% searching for reference images with known camera locations. identifying paired images for further camera pose estimation.
% providing prior, such as raw locations for further camera localization of query images.% and 
Existing image-based methods~\cite{arandjelovic2016netvlad,ge2020self} take an image as input and learn a global description of the image for retrieval. We introduce to refine the retrieval results using correspondences as re-ranking of the image-based approaches.
We study the benchmark of the famous NetVLAD~\cite{arandjelovic2016netvlad} and its follow-up works~\cite{ge2020self, kim2017learned, liu2019stochastic}.
% They generally take an image as input and learn a global description of the image for retrieval.
% We further tackle the task of image-based localization, which aims at localizing a query image by retrieving nearby reference images with geographical tags~\cite{ge2020self}. To this end, we adopt the setting used in~\cite{ge2020self,arandjelovic2016netvlad, kim2017learned, liu2019stochastic}. Note that the task here focuses on predicting \textit{GPS locations} of query images in \textit{large scale} scenes, instead of camera pose estimation~\cite{sattler2016large, taira2018inloc, brachmann2020visual}. Existing image-based methods~\cite{arandjelovic2016netvlad,ge2020self} take an image as input and learn a global description of the image for retrieval.
Here, we achieve the coarse-to-fine retrieval 
% by exploring the applications of correspondences as post-processing 
with the following three steps:
% utilize correspondences to post-process the results of image retrieval and achieve coarse-to-fine retrieval in the following three steps:
1) Use image-based methods~\cite{arandjelovic2016netvlad,ge2020self} to search for the top-$k$ ($k$ is empirically set to 100) images for each query. We did not use all reference images for re-ranking due to the computational cost.
2) Perform feature matching, with either an existing method~\cite{fischler1981random,moo2018learning} or our approach, on each query-retrieved image pair.
3) Re-rank the top-$k$ images using a refined similarity measure defined as $S_\text{img} + S_\text{inl}$, where $S_\text{img}$ represents the original similarity estimated by image-based methods, and $S_\text{inl}$ is the number of selected inliers normalized within $[0, 1]$.
This approach leveraged the fact that image-based methods exploit global description while correspondence-based ones focus on local patterns, making these two kinds of methods complementary. 
Note that the correspondences can be further used to estimate camera pose of query images~\cite{germain2020s2dnet}, but we concentrates on accurate image retrieval in this section.

In our experiments, we use the state-of-the-art SFRS~\cite{ge2020self} as image-based method and employ RANSAC~\cite{fischler1981random}, PointCN~\cite{moo2018learning} or our CLNet as the post-processing technique.
Note that PointCN and CLNet were pretrained on YFCC100M~\cite{thomee2016yfcc100m}.
% we combine CLNet pretrained on YFCC100M with SFRS~\cite{ge2020self} which achieves the best performance reported in~\cite{ge2020self}; the reranking is performed based on $S_f = S_{s} + S_{c}$, where $S_f$ denotes the final score, $S_{s}$ represents the similarity score estimated by SFRS, and $S_{c}$ is the inlier number normalized into $[0, 1]$. 
% We did not rank all reference images by feature-based methods due to the intractable time consuming. 
% Experiments are conducted on Tokyo 24/7 \cite{}, using the same evaluation metric as~\cite{ge2020self}, 
As shown in Table~\ref{tab:retrieval},
CLNet improves SFRS by a considerable margin of $6.0$ percentage points in terms of Recall@1.
% compared with original SFRS, CLNet-SFRS works noticeably better, improving R@1 by $6.0\%$. 
% The complementary relationship between the two kinds of methods is sufficiently leveraged by our combination solution. 
The superiority of our method can be also observed by comparing ``CLNet-SFRS'' with ``RANSAC-SFRS'' and ``PointCN-SFRS'', with our CLNet yielding the best results.

\begin{table}[t]
\scriptsize
    \begin{center}
        \begin{tabular}{l|c|c|c}
        \hline
        \multirow{2}*{Method} & \multicolumn{3}{c}{Tokyo 24/7~\cite{torii201524} (\%)} \\
        \cline{2-4} & R@1 & R@5 & R@10 \\
        \hline
        NetVLAD~\cite{arandjelovic2016netvlad} & 73.3 & 82.9 & 86.0 \\
        CRN~\cite{kim2017learned} & 75.2 & 83.8 & 87.3 \\
        SARE~\cite{liu2019stochastic}& 79.7 &  86.7 & 90.5 \\
        SFRS~\cite{ge2020self} & 85.4 & 91.1 & 93.3 \\
        \hline
        RANSAC~\cite{fischler1981random}-SFRS & 88.6 (+3.2) & 93.0 (+1.9) & 93.7 (+0.4) \\
        PointCN~\cite{moo2018learning}-SFRS & 89.5 (+4.1) & 93.3 (+2.2) & 94.3 (+1.0) \\
        Our CLNet-SFRS & \textbf{91.4 (+6.0)} & \textbf{94.0 (+2.9)} & \textbf{94.3 (+1.0)} \\
        \hline
        \end{tabular}
    \end{center}
    \caption{\textbf{Evaluation on the image localization dataset Tokyo 24/7~\cite{torii201524}.} We report the Recall@1/5/10.} %Correspondence-based methods \cite{fischler1981random,moo2018learning} perform as post-processing techniques for image-based methods \cite{ge2020self}.}
    % The recall of retrieved results under different thresholds, i.e., top-1 (R@1), top-5 (R@5), and top-10 (R@10) is shown.}
    \label{tab:retrieval}
    \vspace{-6pt}
\end{table}

\subsection{Ablation Studies}
\label{sec:analysis}
\begin{figure}
    \centering
    \subfigure[]
	{ \includegraphics[width=0.46\linewidth]{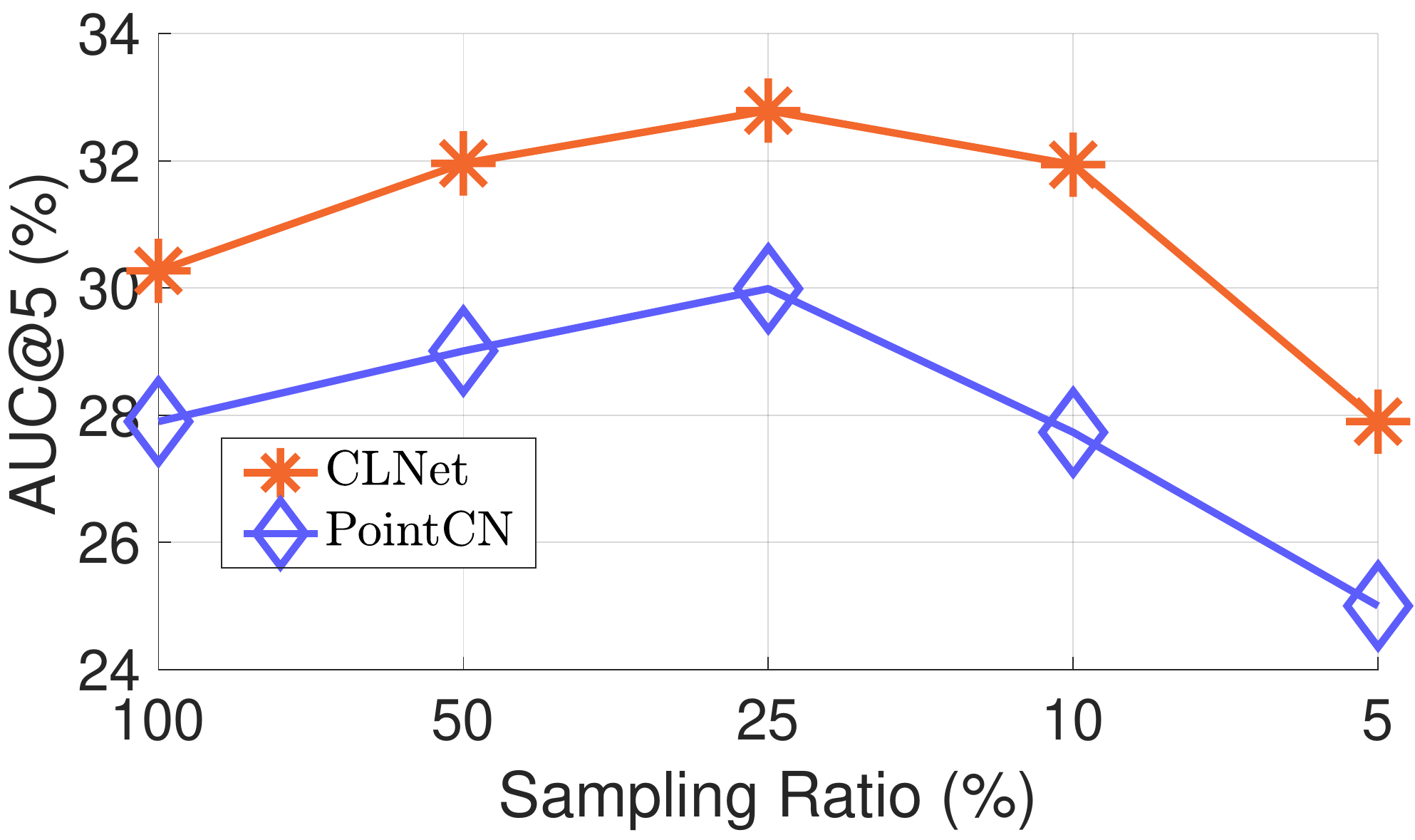}\label{fig:rate}}
	\subfigure[]
	{ \includegraphics[width=0.46\linewidth]{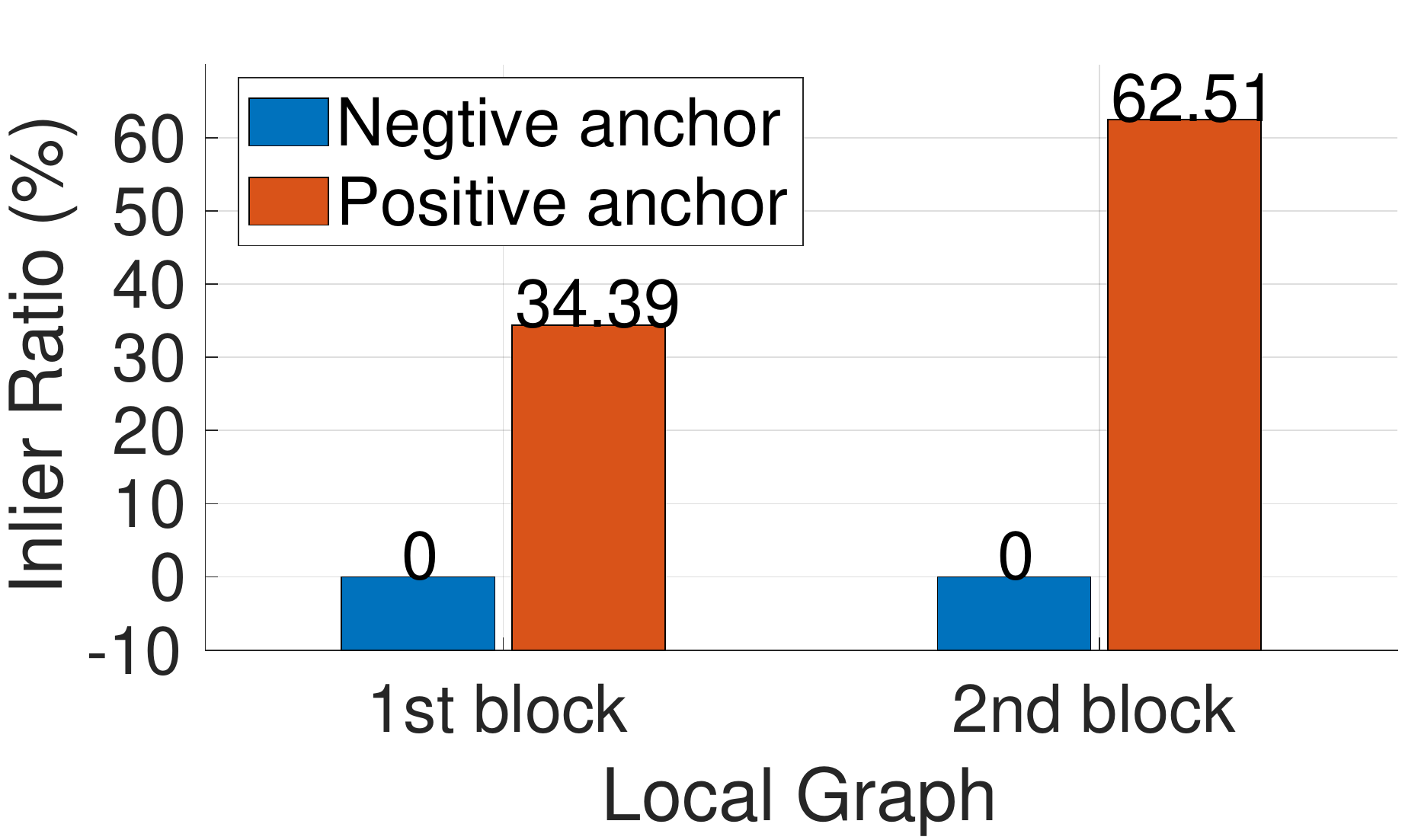}\label{fig:consensus} }	
   % {\includegraphics[width=1.0\linewidth]{figures/sampling_rate.pdf}}
   \caption{\textbf{Effectiveness of our consensus-guided pruning.} (a) AUC@$5^{\circ}$(\%) estimated from candidates sampled by CLNet and PointCN~\cite{moo2018learning} with varying sampling ratios; (b) Inlier ratios of the nodes in the local graphs of CLNet anchored on inliers (positive anchor) and outliers (negative anchor).}
    \label{fig:sampling}
    \vspace{-10pt}
\end{figure}

%\begin{figure}
%    \centering
%    {\includegraphics[width=1.0\linewidth]{figures/inlier_ratio.pdf}}
%   \caption{The inlier ratio (blue line) of candidates and the estimated mAP5 (orange line) with varying down-sampling iterations on YFCC100M are shown.}
%\label{fig:inlier_ratio}
%\end{figure}
\begin{figure}[t]
    \centering
    {\includegraphics[width=0.55\linewidth]{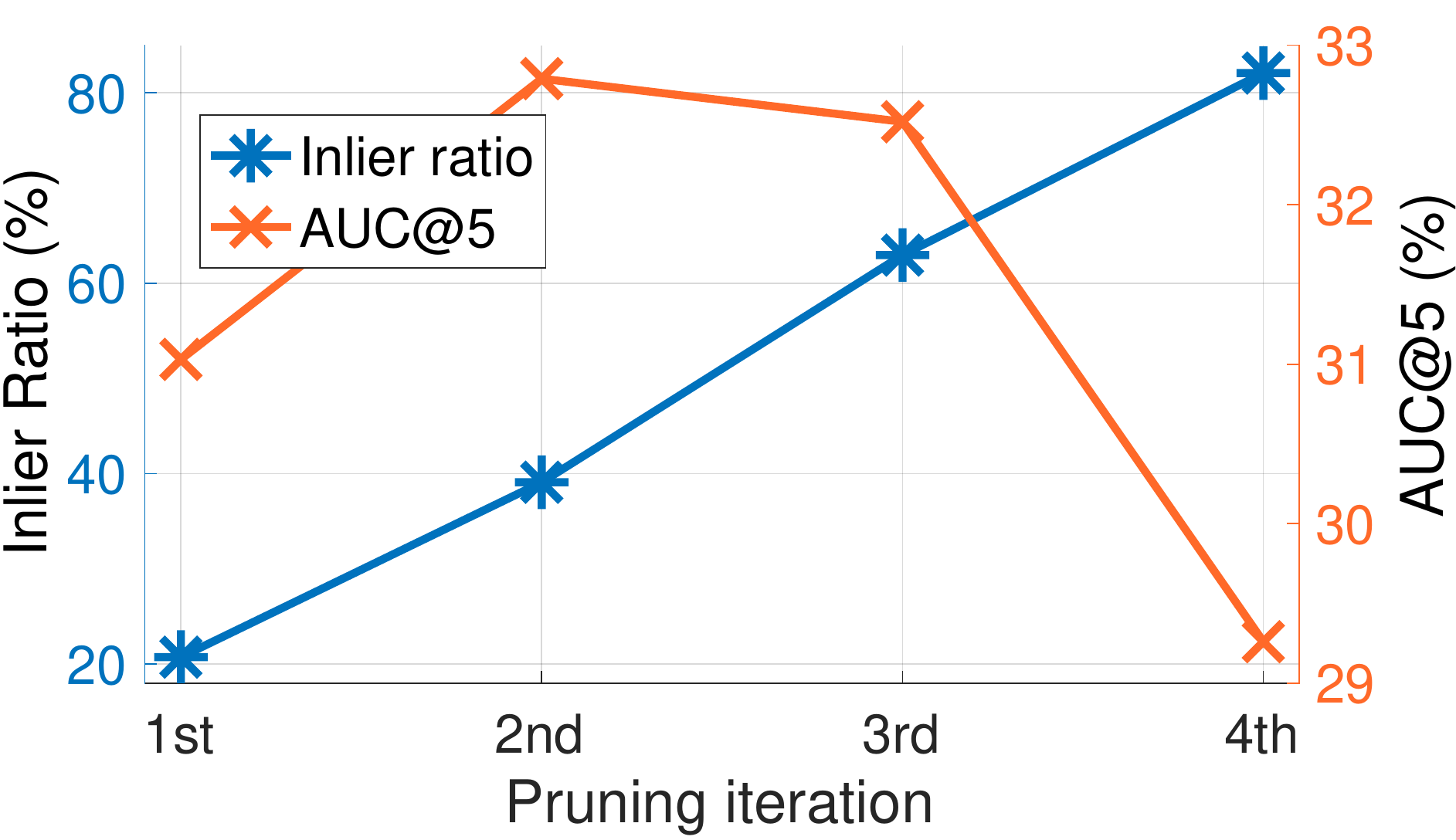}}
   \caption{\textbf{Influence of the number of pruning iterations.} We show the inlier ratio of pruned candidates (blue line) and the AUC@$5^{\circ}$(\%) (orange line) for different numbers of pruning blocks. }
   \vspace{-6pt}
\label{fig:ratio}
\end{figure}

\paragraph{Consensus-guided pruning.}
An intuitive solution to acquire candidates from initial data could consist of sampling matches according to the weights predicted by off-the-shelf methods, \textit{e.g.}, PointCN~\cite{moo2018learning}. We therefore evaluate the benefits of our consensus learning strategy over this baseline.
% replace the consensus guided sampling with PointCN as a comparison. 
Specifically, as baseline, we iteratively perform PointCN twice, which progressively prunes putative correspondences into candidates with a specific sampling ratio; inliers are then identified from the candidates and used to estimate the essential matrix. 
Fig.~\ref{fig:rate} compares the baseline to our method with different sampling ratios on YFCC100M. Our CLNet consistently surpasses the baseline by a large margin, which demonstrates the superiority of our consensus-guided pruning strategy. The best AUC of our CLNet is observed with a sampling ratio of $25\%$, \textit{i.e.}, $25\%$ correspondences are sampled as candidates. In Fig.~\ref{fig:consensus} we further explain the effectiveness of our consensus learning by providing inlier ratios of the nodes in the local graphs of our CLNet. The grouped neighbors contain more inliers for graphs anchored on inliers than for graphs anchored on outliers. This demonstrates that our method is capable of enlarging the correspondence consensus in inlier-anchored graphs, while decreasing the correspondence consensus in outlier-anchored ones. %\MS{Is the last statement really true? Their is still a consensus for outlier-anchored graphs, but the consensus is that they contain outliers.}.

\begin{figure}[t]
    \begin{center}
        {\includegraphics[width=1.0\linewidth]{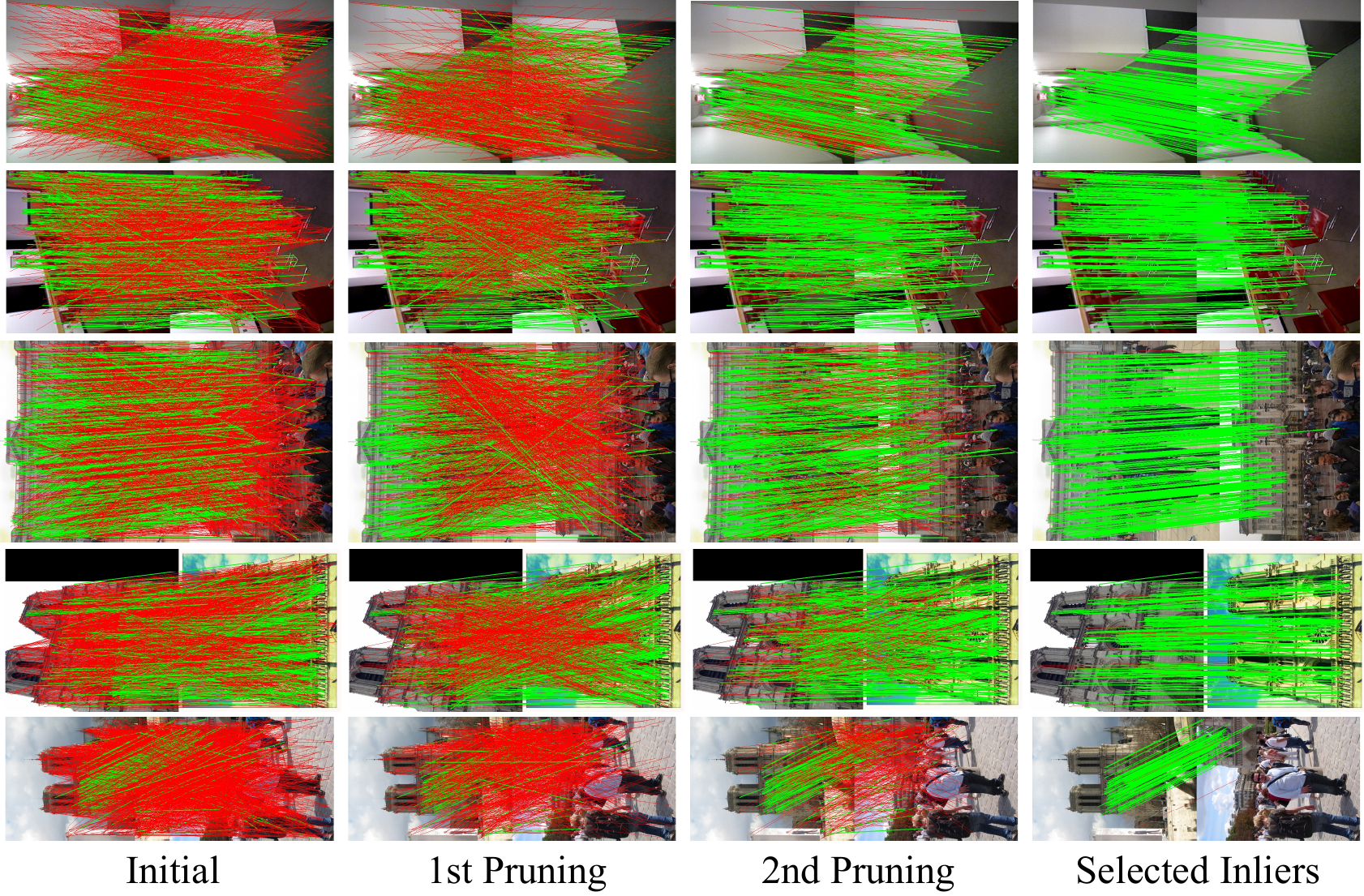}}
    \end{center}
  %  \vspace{-5pt}
   \caption{\textbf{Visualization of progressive pruning}. Outliers and inliers are indicated by red lines and green lines, respectively.}
  % \vspace{-3pt}
\label{fig:visual}
\end{figure}

Since we achieve pruning in an iterative fashion, by stacking pruning blocks, we analyze the effect of the number of pruning iterations on candidate consistency and pose estimation accuracy. As shown in Fig.~\ref{fig:ratio}, iterative pruning yields an increase of inlier ratio, \textit{i.e.}, from 20\% to 80\%, which indicates that the candidates are increasingly more consistent as more pruning iterations are performed. The AUC drops after the second iteration, because the number of remaining matches is too small to carry out robust model estimation. Some visual results are shown in Fig.~\ref{fig:visual}. The imbalance of initial matches is alleviated by progressively removing the outliers, making it increasingly easier to identify the inliers.
%\vspace{-8pt}
%\paragraph{Component analysis.}

\begin{table}[t]
\footnotesize
    \begin{center}
        \begin{tabular}{l|c|c}
            \hline
             & Annular Conv. & MLP \& Max-pooling \\
            \hline
            AUC@$5^{\circ}$ (\%) & \bf{32.79} & 31.99 \\
            \hline
        \end{tabular}
    \end{center}
    \caption{\textbf{Comparison between the proposed annular convolution and an mlp-pooling strategy.} ``MLP \& Max-pooling'' extracts and aggregates local features with MLPs and max-pooling layers, respectively.
    % Conv. represents the convolution processing. 
    % Conv. \& Max-pooling means extracting and aggregating local features by convolution and max-pooling layers, respectively. 
    We report the AUC@$5^{\circ}$(\%) on YFCC100M~\cite{thomee2016yfcc100m}.
    }
    \label{tab:annular}
    \vspace{-6pt}
\end{table}

\begin{table}[t]
    \footnotesize
    \begin{center}
        \begin{tabular}{c|c|c|c}
            \hline
            Local Cons. & Global Cons. & Adaptive temp. & AUC@$5^{\circ}$ (\%) \\
            \hline
            \xmark & \xmark & \xmark & 29.99 \\
            \cmark & \xmark & \xmark & 30.94 \\
            \cmark & \cmark & \xmark & 31.70 \\
            \cmark & \cmark & \cmark & 32.79 \\
            \hline
        \end{tabular}
    \end{center}
    \caption{\textbf{Influence of the individual components of CLNet.} All models were trained and tested on YFCC100M~\cite{torii201524}.}
    \label{tab:ablation}
  %  \vspace{-10pt}
\end{table}

To further understand the importance of each component in CLNet, in Table~\ref{tab:annular}, we compare the proposed annular convolution with the mlp-pooling strategy that has been employed in~\cite{qi2017pointnet++,wang2019dynamic}. Our annular convolution yields a $0.80$ percentage point AUC improvement on YFCC100M, demonstrating its effectiveness. Furthermore, we evaluate different combinations of our method's components in Table~\ref{tab:ablation}. As expected, all components contribute to the optimal performance of our approach. In particular, local-to-global consensus learning leads to a $1.71$ percentage point AUC improvement (third row \textit{vs.} first row), and the performance is further boosted by applying adaptive temperatures. Note that progressive pruning is utilized in all cases.

\section{Conclusion}

To overcome the negative impact of the dominant outliers, we have proposed to progressively prune the putative correspondences into more reliable candidates with a local-to-global consensus learning network. Our framework builds local and global graphs on-the-fly, which \textit{explicitly} describe the correspondence consensus in local and global contexts, facilitating  pruning. Our experiments in diverse scenarios have demonstrated that the proposed progressive pruning strategy largely alleviates the effect of randomly distributed outliers, showing significant improvements over state-of-the-arts on multiple benchmarks. As the candidates are sampled using a constant ratio in our current framework, we will consider about a pruning strategy with adaptive ratios in our future work.
%\MS{Some suggestions for future work?}
%The most confident inliers selected from the denoised subset are used for robust model estimation. Our proposed correspondence denoising largely alleviates the randomly distributed outliers in various scenarios, showing significant improvements on multiple benchmarks.

{\small
\bibliographystyle{ieee_fullname}
\bibliography{egbib}
}

\end{document}